\def\year{2022}\relax
\documentclass[letterpaper]{article} %

 \PassOptionsToPackage{numbers, compress}{natbib}
 
\usepackage{aaai22}  %
\usepackage{times}  %
\usepackage{helvet}  %
\usepackage{courier}  %
\usepackage[hyphens]{url}  %
\usepackage{graphicx} %
\def\UrlFont{\rm}  %
\usepackage{natbib}  %
\usepackage{caption} %
\DeclareCaptionStyle{ruled}{labelfont=normalfont,labelsep=colon,strut=off} %
\usepackage{algorithm}
\usepackage{algorithmic}

\usepackage{amsfonts}
\usepackage{graphicx}
\usepackage{amsmath}
\usepackage{cleveref}

\usepackage{mdframed}

\newcommand{\be}{\begin{equation}}
\newcommand{\ee}{\end{equation}}
\definecolor{Gray}{gray}{0.85}
\definecolor{LightCyan}{rgb}{0.88,1,1}
\usepackage{xcolor}

\makeatletter
\usepackage{xspace} 
\def\@onedot{\ifx\@let@token.\else.\null\fi\xspace}
\DeclareRobustCommand\onedot{\futurelet\@let@token\@onedot}

\definecolor{blue1}{RGB}{0,128,255}
\definecolor{blue3}{RGB}{0,0,128}
\definecolor{darkpastelgreen}{rgb}{0.01, 0.75, 0.24}
\definecolor{cerulean}{rgb}{0.0, 0.48, 0.65}

\def\eg{\emph{e.g}\onedot}

\def\ie{\emph{i.e}\onedot}

\def\eqref#1{equation~\ref{#1}}

\def\1{\bm{1}}

\def\rvx{{\mathbf{x}}}

\DeclareMathAlphabet{\mathsfit}{\encodingdefault}{\sfdefault}{m}{sl}
\SetMathAlphabet{\mathsfit}{bold}{\encodingdefault}{\sfdefault}{bx}{n}

\newcommand{\x}{\ensuremath{\mathbf{x}}\xspace}

\DeclareMathOperator*{\argmin}{arg\,min}

\newtheorem{thm}{Theorem}[section]

\usepackage{amssymb}
\usepackage{array}
\usepackage{multirow}
\usepackage{booktabs}
\usepackage{color, colortbl}
\usepackage{makecell}
\usepackage{float}
\usepackage{thm-restate}
\usepackage{diagbox}

\usepackage{newfloat}
\usepackage{listings}
\floatstyle{ruled}
\newfloat{listing}{tb}{lst}{}
\floatname{listing}{Listing}
\newcommand{\model}[1]{\textsc{IS-Count}}

\title{\model{}: Large-scale Object Counting from Satellite Images \\ with Covariate-based Importance Sampling}

\title{\model{}: Large-scale Object Counting from Satellite Images \\ with Covariate-based Importance Sampling}
\author {
    Chenlin Meng\footnote{Joint first authors.}, Enci Liu\textsuperscript{*},
    Willie Neiswanger, Jiaming Song, \\Marshall Burke, David Lobell, Stefano Ermon
}
\affiliations {
    \vspace{1mm}
    Stanford University\\
    \vspace{2mm}
    \{chenlin, jesslec, neiswanger, tsong\}@cs.stanford.edu,
    \{mburke, dlobell\}@stanford.edu,
    ermon@cs.stanford.edu
}

\begin{document}

\maketitle

\begin{abstract}

Object detection in high-resolution satellite imagery is emerging as a scalable alternative to on-the-ground survey data collection in many environmental and socioeconomic monitoring applications. 
However, 
performing object detection over large geographies can 
still 
be prohibitively expensive due to the high cost of purchasing imagery and compute. Inspired 
by 
traditional survey data collection strategies, we propose an approach to estimate object count statistics over large geographies through sampling. 
Given a cost budget, our method selects a small number of representative areas
by sampling from a learnable proposal distribution. 
Using importance sampling, 
we are able to accurately estimate object counts after processing only a small fraction of the  images compared to an exhaustive approach. 
We show empirically that the proposed framework achieves strong performance on estimating the number of buildings in the United States and Africa, cars in Kenya, brick kilns in Bangladesh, and swimming pools in the U.S., while 
requiring as few as
0.01\% of satellite images compared to an exhaustive approach.

\end{abstract}
\section{Introduction}
\begin{figure*}
    \centering
    \includegraphics[width=1.0\textwidth]{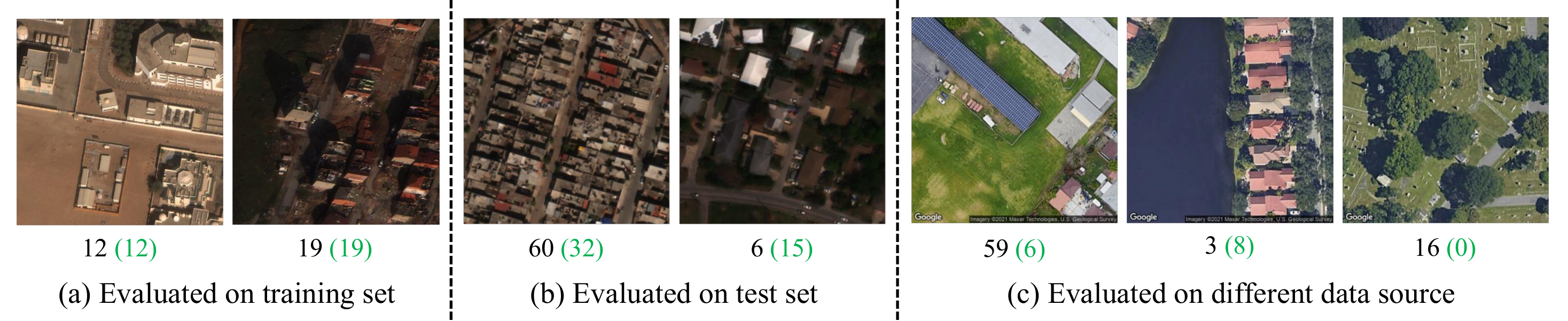}
    \caption{Predicted building counts of a YOLO-v3 model \cite{redmon2017yolo9000} trained on xView \cite{lam2018xview}. Ground truth building counts are highlighted in green and provided in the parenthesis next to the model prediction. We observe that the model has an almost perfect prediction on the training set (see Figure (a)). However, the performance degrades on the test set (see Figure (b)) and satellite imagery from other data sources (see Figure (c)). Additional training details of the object counting model are provided in Appendix.
    }
    \label{fig:covariate_shift_failure}
\end{figure*}

The quantity and location of human-made objects are key information for the measurement and understanding of human activity and economic livelihoods. 
Such physical capital---for instance, buildings, cars, and roads---is both an important component of current economic well-being as well as an input into future prosperity. Information on physical capital has traditionally been derived from ground-based surveys of households, firms, or communities~\cite{bea2003consumer}.  However, because such surveys are expensive and time consuming to conduct, key data on physical capital and related livelihood measures are lacking for much of the world, inhibiting our understanding of the patterns and determinants of economic activity~\cite{burke2021using}.

Object detection in high-resolution satellite imagery has emerged as a scalable alternative to traditional survey-based approaches to gathering data on economic activity.  For instance, imagery-based counts of the number of buildings at country level allows policymakers to monitor progress towards economic development~\cite{Ayush2021EfficientPM,uzkent2020efficient,Uzkent_2020,sheehan2019predicting,blumenstock2015predicting,Yeh2020UsingPA},
counts of the number of brick kilns allows environmental scientists to track pollution from informal industries~\cite{lee2021scalable}, counts of multiple objects enables accurate poverty prediction~\cite{jean2016combining,Ayush2021EfficientPM}, and counts of solar panels in high-resolution imagery enables understanding of green energy adoption at broad scale~\cite{yu2018deepsolar}.

However, due to the substantial cost of purchasing high-resolution imagery, and the large amount of computation needed to estimate or apply models at scale, performing object detection over large geographies is often prohibitively expensive~\cite{uzkent2019learning,Uzkent_2020}, especially if estimates need to be updated. For instance, at standard pricing for high-resolution imagery, purchasing country-wide imagery for one year would cost roughly \$3 million for Uganda, \$15 million for Tanzania, and \$38 million in the Democratic Republic of Congo.\footnote{We assume a price per sq km of \$17 for 3-band imagery, consistent with current industry rates.} Such costs inhibit the widespread application and adoption of satellite-based approaches for livelihood measurement.

Here we propose an importance-sampling approach to efficiently generate object count statistics over large geographies, and validate the approach across multiple continents and object types. Our approach draws inspiration from traditional approaches to large-scale ground survey data collection, which use information from prior surveys (\eg, a prior census) to draw sample locations with probability proportionate to some covariate of interest (\eg, village population). %
In our setting, we expect most objects of interest (\eg, cars) to have close to zero density in certain regions (\eg forested areas).
In this case, 
sampling locations uniformly at random (\ie, with a uniform proposal)
would have a high variance and require a large number of samples.
We therefore propose to use importance sampling (IS) to
select locations 
from important regions (\eg regions where the counts of cars are expected to be non-zero) by sampling from a proposal distribution.  While a good proposal can significantly reduce variance, 
the optimal proposal distribution is unknown, often complicated, and object-specific. We therefore propose to learn the proposal distribution by relating the presence (or absence) of objects to widely available covariates such as population density, nightlights, etc.

Given a cost budget, our method, \model{},
selects a small number of informative areas, using an object detector or gold-standard human annotations for object counting.
It largely reduces the number of satellite images as well as human annotations compared to an exhaustive approach used by object detectors in many real-world counting tasks, while achieving a high accuracy.
We show with experiments on counting buildings, cars, brick kilns, and swimming pools in Africa, United States, and Bangladesh, that \model{} is able to achieve an accurate estimation while requiring as few as 0.01\% images compared to an exhaustive approach, thus potentially reducing the cost of generating large-scale object counts by four orders of magnitude. 
 Our code is available at \textcolor{gray}{\url{https://github.com/sustainlab-group/IS-Count}}.

\section{Problem Setup}

\begin{figure}
    \centering
    \includegraphics[width=0.95\linewidth]{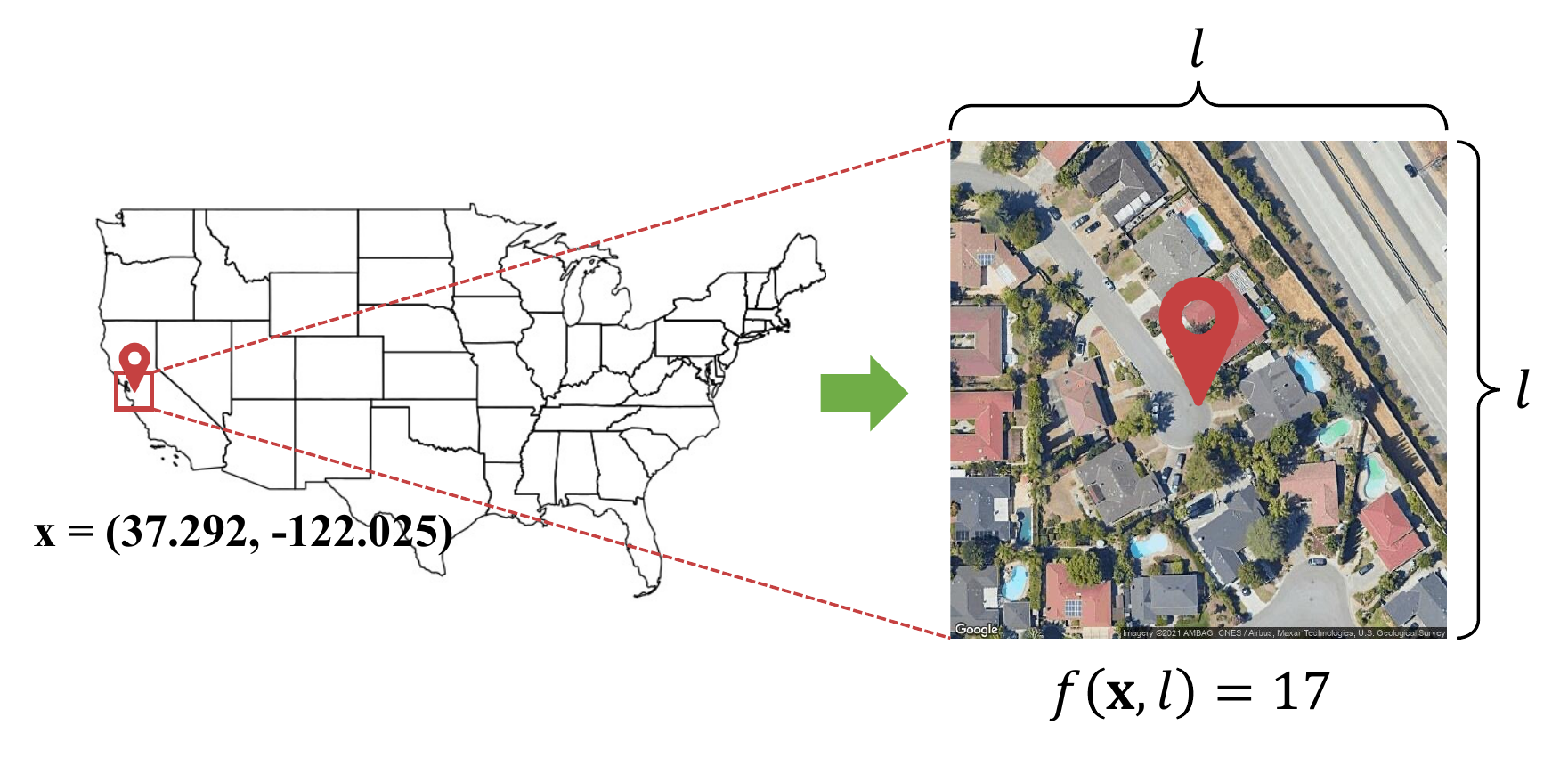}
    \caption{Given a location $\x$, $f(\x,l)$ denotes the total number of objects (e.g. buildings shown here), within the bounding box with size $l \times l$ centered at $\x$.}
    \label{fig:counting_box}
\end{figure}

Given a region $R \subseteq \mathbb{R}^2$ (\eg a country) and a class of target object (\eg buildings) demonstrated with a small number of labeled images, we want to estimate the total number of objects of the desired class that are located within the target region $R$. 
We denote $f(\x, l)$ the total number of objects within a  $l\times l$ bounding box centered at point $\x$ (see \Cref{fig:counting_box}). 
Given a partition of $R$ into non-overlapping $l\times\l$ images with centers $\mathcal{S}$, 
the goal 
is to estimate 
\begin{equation}
    C=\sum_{\x\in \mathcal{S}} f(\x, l).
\end{equation}
While the object count $C$ is less informative than the precise location of all the objects in $R$, object counts are often sufficient in many downstream tasks such as 
regression analyses, e.g. to estimate poverty levels from satellite images~\cite{Ayush2021EfficientPM, ayush2020generating}. Additionally, we will demonstrate that object counts can be obtained much more efficiently, allowing us to scale over large regions where exhaustive mapping would be infeasible.

A naive solution 
to estimate $C$ is to acquire all image tiles covering the entire region $R$, identify the objects in each image (\eg, using a machine learning model), and sum the counts~\cite{crowther2015mapping,yu2018deepsolar,noauthor_scaling_nodate}.
However, this approach has the following limitations.

\paragraph{Satellite Images are Costly}
In many applications,
we need to use high-resolution images for the target objects (\eg, cars) to be visible.
Although low-resolution images are publicly available, high-resolution images typically need to be purchased from private vendors. For instance, it costs approximately 
\$164 million for purchasing 
high-resolution satellite images that cover the entire United States\footnote{\url{https://g3f3z9p8.rocketcdn.me/wp-content/uploads/2018/04/landinfo.com-LAND_INFO_Satellite_Imagery_Pricing.pdf}}
,
which is often infeasible for researchers.

\paragraph{Labeling is Expensive}
If the target region is large, it would be impractical for human annotators to manually count objects in all images. For instance, we estimate it would take approximately 115,000 hours for human annotators to count the total number of cars in Kenya using satellite images\footnote{Estimation based on Amazon Mechanical Turk} (see Appendix).
A more efficient approach is to use an algorithm (typically a machine learning model) to estimate counts~\cite{gao2020counting,bazi2009automatic} or detect objects~\cite{crowther2015mapping,yu2018deepsolar,salami2019fly,mubin2019young,noauthor_scaling_nodate} within each image. However, training such a model often requires very large amounts (\eg 360k) of labeled data \cite{yu2018deepsolar}, whose labels eventually come from human annotators.
As the distribution of satellite images often changes drastically across data sources (\eg Digital Globe and Google Static Map), objects of interest (\eg buildings and farmland), and regions (\eg U.S. and Africa), an object detector pre-trained on one dataset could fail easily on another due to covariate shifts~\cite{noauthor_ai_nodate}  (see \Cref{fig:covariate_shift_failure}).
This makes it hard to directly apply a pre-trained object detector to a new task where sources of satellite images are different even if large-scale labeled datasets are available for the object of interest.

\paragraph{Sampling}
Denote $S_R$ the area of $R$, and $U(R)$ the uniform distribution on $R$.
When $l\ll R$, the total number of objects of interest $C$ in the region $R$ can be computed as
\begin{align}
\label{eq:uniform_sampling}
    C=\frac{S_R}{l^2}\mathbb{E}_{\x\sim U(R)}[f(\x, l)].
\end{align}
The following unbiased estimator is often used to evaluate \Cref{eq:uniform_sampling}
\begin{align}
    \hat{C}=\frac{S_R}{l^2}\frac{1}{n}\sum_{i=1}^{n}f(\x_i, l),\; \x_i\sim U(R),
\end{align}
where $\{\x_i\}_{i=1}^{n}$ are i.i.d. samples from the uniform distribution $U(R)$. 
This can drastically reduce cost if $n$ is small, even allowing for (gold-standard) human annotations for evaluating $f(\x_i, l)$.

In real-world applications, however, it is expected that the object of interest (\eg, buildings) can have a close to zero density in certain regions (\eg, forest).
In this case, estimating object counts directly via uniform sampling would have a high variance and thus require a huge number of samples as we will show in the experimental section.

\section{\model{}: Large-scale Object Counting with Importance Sampling}
\begin{figure*}
    \centering
    \includegraphics[width=1.0\textwidth]{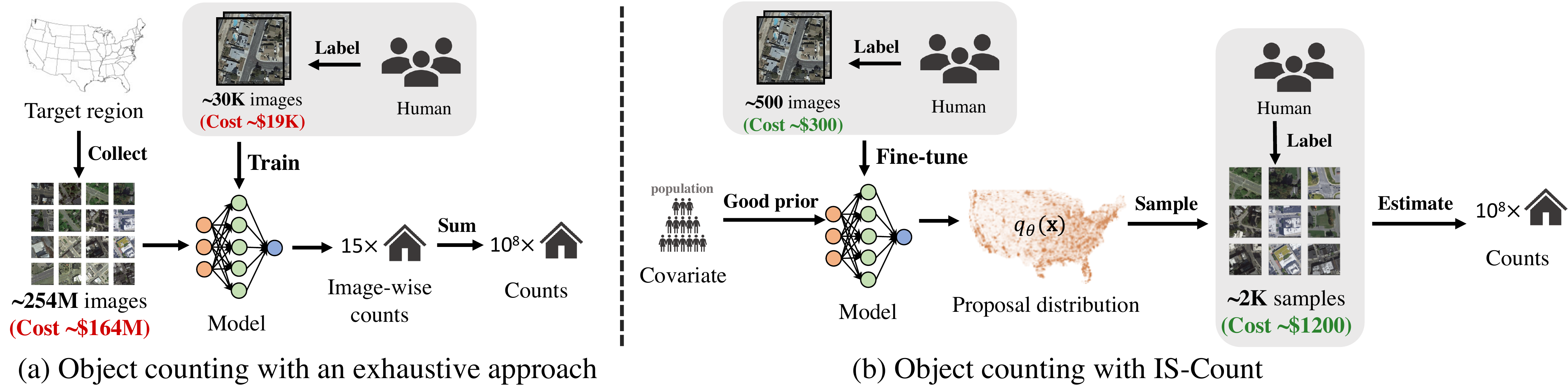}
    \caption{
    Object counting frameworks comparison (example on the US.). Figure (a): An exhaustive approach downloads all image tiles covering the target region, maps the objects in each image using a trained model, and takes the summation of counts in all the images to produce a total count. However, purchasing satellite imagery for a \textbf{large} target region can be expensive. 
    Figure (b): In contrast, \model{}  selects  \textbf{a small  number}  of  informative  areas  for  object  counting  by sampling from a learnable proposal distribution which captures the representative areas. \model{} largely reduces the number  of  satellite  images  and human  annotations while achieving a high accuracy.
    }
    \label{fig:object_count_demo}
\end{figure*}

In this paper, we alleviate the above challenges by proposing an efficient object counting framework that incorporates prior knowledge from socioeconomic indicators into importance sampling (IS). 
Our method, \model{}, provides an unbiased estimation for object counts and requires as few as 0.01\% high-resolution satellite images compared to an exhaustive approach (see \Cref{fig:object_count_demo}). \model{} can be easily adapted to different target regions or object classes using a small number of labeled images, 
while achieving strong empirical performance.

\subsection{Importance Sampling}
Importance sampling (IS) introduces a proposal distribution to choose representative samples from important regions (\eg, regions where the counts are non-zero) with the goal of reducing the variance of uniform sampling. 
Given 
a proposal distribution $q(\x)$ with a full support on $R$, we can estimate the total object count using importance sampling
\begin{align}
\label{eq:is_object_count}
    C=\frac{1}{l^2}\mathbb{E}_{\x\sim q(\x)}\left[\frac{f(\x, l)}{q(\x)}\right].
\end{align}
\Cref{eq:is_object_count} can be approximated by the following unbiased estimator
\begin{align}
\label{eq:is_object_count_n}
    \hat{C}_n=\frac{1}{l^2}\frac{1}{n}\sum_{i=1}^{n}\frac{f(\x_i, l)}{q(\x_i)}, \: \x_i\sim q(\x),
\end{align}
where $\{\x_i\}_{i=1}^{N}$ are i.i.d. samples from $q(\x)$.
The optimal proposal distribution $q^{*}(\x)$ which has the smallest variance
 should be proportional to $f(\x,l)$~\cite{owen2013monte}.

We therefore want to design a proposal distribution that is as close as possible to the object density.
Although high-resolution images are costly, socioeconomic covariates such as nightlight intensity are globally available, free of charge, and correlate strongly human activities~\cite{jean2016combining}.

In the following, we assume that we always have access to certain covariates that are cheap and publicly available for the target region. We treat the covariate as the base distribution for designing the proposal distribution $q(\x)$. In order for the base distribution to capture information specialized to the task, we propose to fine-tune the base distribution using a small number of labeled satellite images, where the labels are count statistics (see \Cref{fig:object_count_demo}). We also provide the pseudocode for the framework in \Cref{alg:algorithm}.

\begin{algorithm}[tb]
\caption{Object counting with \model{}}
\label{alg:algorithm}
\textbf{Input}: Region $R$, object class, budget $n$, covariate,\\ and a small number of labeled examples\\
\textbf{Output}: Estimated object count
\begin{algorithmic}[1] %
\STATE  $q(\x)$ $\leftarrow$ covariate distribution
\STATE Fine-tune $q(\x)$ with labeled examples
\STATE Sample $\{\x_i\}_{i=1}^{n}$ from $q(\x)$
\STATE $\hat{C}_n\leftarrow$ Estimate $C$ using \Cref{eq:is_object_count_n}
\STATE \textbf{return} $\hat{C}_n$
\end{algorithmic}
\end{algorithm}

Our key insight is that the base covariate distribution can provide good prior knowledge for a given task, and therefore we only need a small number of labeled images for fine-tuning to obtain a task-specific proposal distribution that reduces the variance for sampling. 
As this framework only requires a small amount of labeled images for each task and always provides an unbiased estimate, it can be easily adapted to different counting tasks, providing a general framework for large-scale counting.

\subsection{Proposal Distributions with Task-specific Tuning}
Given the base covariate distributions, we can fine-tune the proposal distribution using a small number of labeled satellite images to design a task-specific proposal distribution.
Isotonic regression provides an approach to learning a non-decreasing transformation, allowing fine-tuning the proposal distribution based on the input covariate distribution. 
More specially, 
let $h(\x)\in \mathbb{R}$ be the covariate pixel value at geolocation $\x$, and $f(\x,l)$ be the object count (see \Cref{fig:counting_box}), we learn a non-decreasing map $g_{\theta}(\cdot, l):\mathbb{R}\to \mathbb{R}$ 
which maps $h(\x)$ to be close 
to its corresponding object count $f(\x, l)$. The objective function is defined as
\begin{equation}
\label{eq:isotonic_regression}
    \theta=\argmin_{\theta} \sum_{i=1}^{n}w_i(g_{\theta}(h(\x_i), l))-f(\x_i,l))^2,
\end{equation}
subject to $g_{\theta}(a,l)\le g_{\theta}(b,l)$ whenever $a\le b$. 
In \Cref{eq:isotonic_regression}, $w_i$ are positive weights
and $\{\x_i\}_{i=1}^{n}$ are coordinates sampled from the target region $R$. 

Although it might seem natural to use a general regression model, which takes multiple input covariates, to predict the corresponding object count, we observe that such approach requires a much larger number of training data to learn a robust model. As the training size is deducted from the sampling budget, a general regression does not have a strong estimation performance due to the reduction in sample size. We observe that isotonic regression, being more restrictive, has stronger empirical performance than general regressions when the training size is limited, potentially due to the strong inductive bias of being monotonic so that a higher covariate value would be mapped to a higher object density. 

\begin{figure*}
    \centering
    \includegraphics[width=1.0\textwidth]{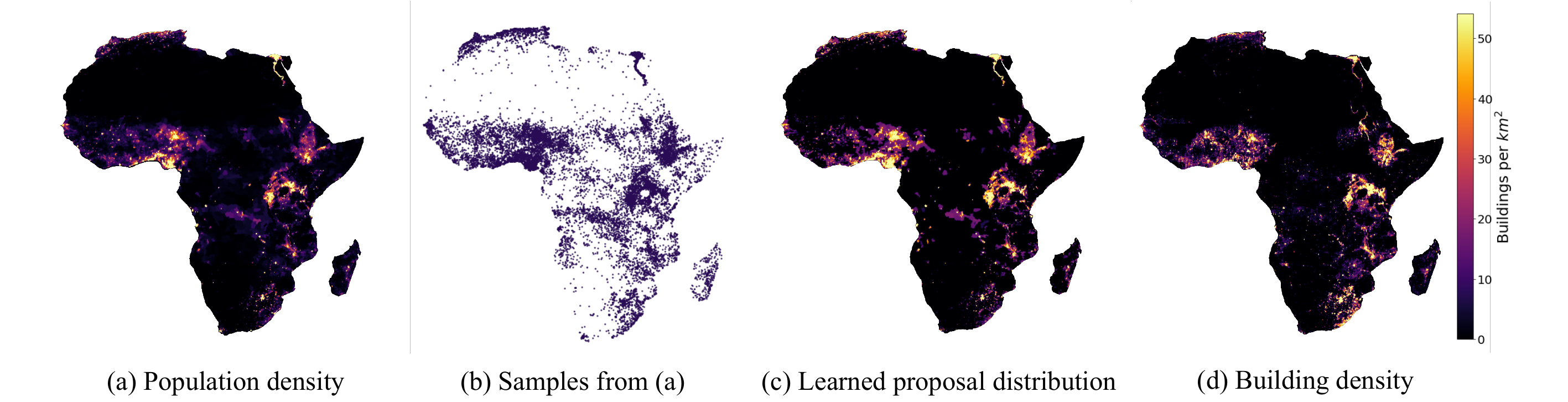}
    \caption{Visualization of distributions in Africa. Figure (a) shows the population density in Africa; Figure (b) shows samples drawn from the population density shown in (a); Figure (c) shows the proposal distribution learned by isotonic regression with (a) as inputs; Figure (d) shows the building density derived from the Google Open Buildings dataset~\cite{Sirko2021ContinentalScaleBD}.}
    \label{fig:dataset}
\end{figure*}

Given the learned model $g_{\theta}(\cdot,l)$, the proposal distribution $q_{\theta}(\x)$ can be derived as
\begin{align}
    q_{\theta}(\x,l) \triangleq \frac{g_{\theta}(h(\x),l)}{\int_{\x}g_{\theta}(h(\x),l) d\x},
\end{align}
where $\int_{\x}g_{\theta}(h(\x),l)d\x$ is computed by taking the summation of all covariate elements weighted by their volumes in the region $R$, which can be achieved because there are only finitely many elements in the covariate dataset.
As a special case, when $g_{\theta}(h(\x))$ is proportional to $h(\x)$ for all $\x\in R$, sampling from $q_{\theta}(\x,l)$ is equivalent to sampling proportionally to the covariate distribution. We provide a visualization of a learned proposal distribution in \Cref{fig:dataset}.

\subsection{Theoretical Analysis}
Given the estimation $\hat{C}_n$ from \Cref{eq:is_object_count_n}, we can bound its estimation error using the Kullback-Leibler (KL) divergence between $q^{*}(\x)$ and $q_{\theta}(\x,l)$. As $l$ is a constant, we use $q_{\theta}(\x)$ to denote $q_{\theta}(\x,l)$ for simplicity. 
\begin{restatable}[]{proposition}{thm}
\label{eq:importance_sampling_bound}
Suppose $\frac{q^{*}(\x)}{q_{\theta}(\x)}$ is well defined on  $R$, let $L=\text{KL}(q^{*}(\x)\Vert q_{\theta}(\x))$, \ie the KL divergence between $q^{*}(\x)$ and $q_{\theta}(\x)$. If $n=\exp(L+t)$ for some $t\ge 0$, then
\begin{equation}
\resizebox{0.95\linewidth}{!}{
    $\mathbb{E}[\vert \hat{C}_n-C\vert ]\le C\bigg(e^{-\frac{t}{4}}+2\sqrt{\mathbb{P}(\log \frac{q^{*}(\x)}{q_{\theta}(\x)}>L+\frac{t}{2})}\bigg)$.
}    
\end{equation}
\end{restatable}
\Cref{eq:importance_sampling_bound} provides a bound for the estimation error of $\hat{C}_n$ using $n$ samples. 
Intuitively, when $p(\x)$ is close to $p^{*}(\x)$, the count estimation using sampling is more precise with limited samples.
When $q_{\theta}(\x)=q^{*}(\x)$ almost everywhere, we have $\text{KL}(q^{*}(\x)\Vert q_{\theta}(\x))=0$, which implies  
\begin{equation}
    \mathbb{E}[\vert \hat{C}_{\exp(t)}-C\vert ]\le Ce^{-\frac{t}{4}}.
\end{equation}
The proof of \Cref{eq:importance_sampling_bound} can be derived from \citeauthor{chatterjee2018sample}, and we provide details in Appendix.
 
 We can also provide a probability lower bound for the estimation $\hat{C}_n$ using Markov inequality.
\begin{restatable}[]{theorem}{inequality}
\label{thm:markov_inequality}
For any $k>0$
\begin{equation}
    \mathbb{P}[\hat{C}_n\ge kC]\le \frac{1}{k}.
\end{equation}
\end{restatable}
\Cref{thm:markov_inequality} implies that the probability that the estimation $\hat{C}_n$ is $k$ times larger than the ground truth (GT) counts $C$ is no more than $\frac{1}{k}$, providing a probability lower bound for the estimation. 
Intuitively, we would want our proposal distribution to satisfy that $\frac{q_{\theta}(\x)}{f(\x,l)}$ has a small variance, so that $q_{\theta}(\x)$ will be close to $q^{*}(\x)$.
In the following, we provide a lower bound for the variance of $\frac{q_{\theta}(\x)}{f(\x,l)}$ based on the KL divergence between $q^{*}$ and $q$.
\begin{restatable}[]{theorem}{variance}
\label{thm:v_bound}
The variance of $\frac{q_{\theta}(\x)}{f(\x,l)}$ is an upper bound to $C^2 (e^{KL(q^{*}(\x) \Vert q_{\theta}(\x))} - 1)$.
\end{restatable}
\Cref{thm:v_bound} implies that the variance of the estimator cannot be too small unless the KL divergence between $q^{*}$ and $q$ is also small, which encourages us to reduce the KL divergence. We provide the proof in Appendix.

\section{Experimental Setup}
\begin{figure*}[h]
    \centering
    \includegraphics[width=1.0\textwidth]{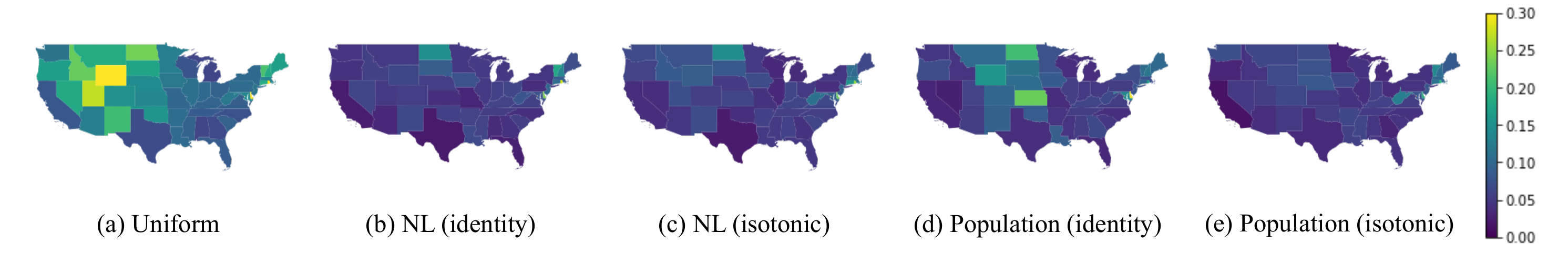}
    \caption{Estimation error (see \Cref{eq:error}) on the building counts in the contiguous United States (the darker the better). We treat the count from the MS Building Footprints as the ground truth to compare with. All results are averaged over 20 runs and using $0.2\%$ images, meaning that the ratio between the total  area covered by the used images in the budget and the area of the target region is $0.2\%$.
    }
    \label{fig:building_us_error}
\end{figure*}

To show the effectiveness, generality, and scalability of \model{}, we conduct experiments on multiple real-world applications.

\subsection{Tasks and Datasets}
We consider the following four tasks across 45 countries: 1) counting buildings in the US and 43 African countries; 2) counting cars in Kenya; 3) counting brick kilns in Bangladesh; and 4) counting swimming pools in the US. 
We emphasize that our approach is generalizable to other counting tasks, 
and the tasks and the regions of interest are chosen such that the regions are reasonably large and validation data are either available or readily obtained from human labeling of images. 
We provide more details in Appendix.
\paragraph{Task 1: Buildings} Building count provides insights into urban development and assists humanitarian efforts~\cite{Sirko2021ContinentalScaleBD}. We
evaluate \model{} on 
counting buildings in the US. and 43 African countries. We use the Microsoft Building Footprints\footnote{\url{https://github.com/microsoft/USBuildingFootprints}}
and the Google Open Buildings \cite{Sirko2021ContinentalScaleBD}
as the ground truth building distribution for the US. and Africa respectively. 
The image-wise building counts are directly extracted from the corresponding datasets.

\paragraph{Task 2: Cars} The number of cars correlates with the economic development of a region, especially in low-income countries~\cite{litman2002automobile,li2020estimation}.
We focus on counting cars in a region roughly covering Nairobi, the capital of Kenya. 
We collect all satellite images covering the entire region, and hand-label the count of cars in all images, the sum of which
is treated as the ground truth.

\paragraph{Task 3: Brick kilns} Brick manufacturing is a major source of pollution in South Asia but is dominated by small-scale producers who are hard to monitor~\cite{lee2021scalable}. Understanding the distribution of brick kilns is thus of importance for policymakers. In this section, we perform experiments on counting brick kilns in Bangladesh.
We use the dataset from \cite{lee2021scalable} and treat their results 
as the ground truth count and collect image-wise counts as for buildings.

\paragraph{Task 4: Swimming pools} The number of swimming pools informs policy makers about urban planning and assists larger-scale water quality control~\cite{noauthor_water_nodate}. In this task, we estimate the count of swimming pools in the US. at country-level. 
It is estimated that there are 10,709,000 swimming pools in the United States~\cite{noauthor_pools_nodate}, 
which we treat as the ground truth. 
As we are not aware of existing datasets on swimming pools, we sample a small amount of images
and collect the counts from human annotators (Amazon MTurk) for estimating the total count.

\subsection{Base Distributions from Covariates}
For constructing base proposal distributions, we focus on two covariates, population density and nightlight intensity, and leave the exploration of other covariates for future work.

\paragraph{Population}
The population density raster is a single-channel image,
with each pixel a positive float value denoting the (estimated) population
for a 1000m$\times$1000m area on the ground (see \Cref{fig:dataset}).
We can derive a density function $q(\x)$ to be proportional to the population density by dividing each pixel value with the normalization constant---the summation of all pixel values in the population raster. We treat the density within each pixel as a uniform distribution. The derived density $q(\x)$ can be used as the proposal distribution for \model{}.
More details can be found in Appendix.

\paragraph{Nightlight} 
We use the global nightlight raster, which is also a single channel image, with each pixel a positive float value denoting the nightlight intensity for a 750m$\times$750m area %
(see Appendix \Cref{app:fig:nl_data}).
Similarly, we can derive a density function $q(\x)$ to be proportional to the nightlight intensity and treat the density within each pixel to be uniform.

\subsection{Experiment Settings}

We consider the following three settings for constructing the proposal distribution for \model{}. 

\paragraph{Identity} We directly use the base covariate distribution as the proposal distribution without learning.

\paragraph{Isotonic}
We fine-tune the base proposal distribution with isotonic regression using a small number of labeled samples (\eg, 100 samples).
We deduct the size of the training data from the total budget, meaning that the larger the training set, the fewer satellite images we can sample for count estimation.

\paragraph{Isotonic$^\star$} 
Depending on task, there could already exist a certain amount of observed labeled data, which could potentially be sampled from an unknown distribution. Although these data might not be used for count estimation as they are not sampled from the proposal distribution, they can still be used to fine-tune the proposal distribution. This observation motivates us to have the second isotonic setting where the size of the training data is not deducted from the total budget.

\subsection{Evaluation Metrics}
We evaluate the performance using percent error
defined as
\begin{equation}
\label{eq:error}
    \text{Error} = \frac{|\hat{C}_{n}-C|}{C}\times 100\%,
\end{equation}
where $C$ is the ``ground truth" (GT) object count obtained from existing datasets or human annotators, and $\hat{C}_{n}$ is the estimation using $n$ samples (see \Cref{eq:is_object_count_n}).

\begin{table}[t]
    \centering
    \caption{Averaged error over all states in the US. and 43 African countries. All results are averaged over 20 runs. The total area of used satellite images covers 0.1\% of each target state (US) and country (Africa). The result of uniform sampling is provided in the parenthesis next to the region name. We use 20\% of the budget for training isotonic regression.
    }
    \resizebox{0.92\linewidth}{!}{
    \begin{tabular}{cccccccc}
    \Xhline{3\arrayrulewidth}
    &\multicolumn{2}{c}{US states (16.77\%)}
    &\multicolumn{2}{c}{African countries (21.02\%)}\\
    \Xhline{2\arrayrulewidth}
    Methods  
    &\multicolumn{1}{c}{NL}
    &\multicolumn{1}{c}{Population}
    &\multicolumn{1}{c}{NL}
    &\multicolumn{1}{c}{Population}\\
     \Xhline{2\arrayrulewidth}
     Identity &8.46\% &9.71\%  %
     &15.58\% &18.27\% \\
     Isotonic &8.80\% &8.08\% &15.79\% &17.43\%\\
     Isotonic$^{*}$ &\textbf{8.09\%} & \textbf{7.48\%}  & \textbf{14.86\%} & \textbf{16.85\%}\\
    \Xhline{3\arrayrulewidth}
    \end{tabular}
    }
    \label{tab:object_count_state_averaged}
\end{table}

\newcolumntype{G}{>{\centering}b{0.065\textwidth}}
\newcolumntype{C}{>{\centering\arraybackslash}b{0.065\textwidth}}
\definecolor{LightCyan}{rgb}{0.88,1,1}
\definecolor{LightOrange}{rgb}{1,0.85,0.70}
\definecolor{DarkCyan}{rgb}{0,0.8,0.8}
\definecolor{DarkOrange}{rgb}{1,0.50,0.0}
\begin{table*}[h]
    \centering
    \caption{Error rate (\%) of object count estimation using different methods (averaged over 20 runs).
    ``Percentage sampled" denotes the ratio between the total  area covered by the used images and the area of the target region. 
    For isotonic based methods, we use 20\% of the total budget as the training size.
    We report the cost and time saved for  purchasing high-resolution images and labeling with human compared to an exhaustive approach on the target region. Details are provided in Appendix.
    }
    \resizebox{\textwidth}{!}{
    \begin{tabular}{cccccccccc}
    \Xhline{3\arrayrulewidth}
    &\multicolumn{2}{c}{Buildings (US)}
    &\multicolumn{2}{c}{Cars (Nairobi region)} &\multicolumn{2}{c}{Brick Kilns (Bangladesh)} &\multicolumn{1}{c}{Swimming Pools (US)}\\
    \Xhline{2\arrayrulewidth}
     Percentage sampled &$0.001\%$ & $0.01\%$ &$2.6\%$ & $3.9\%$ &$2.0\%$ & $4.0\%$ & $0.001\%$\\
    \Xhline{2\arrayrulewidth}
    \multicolumn{1}{c|}{Uniform} & $9.58\pm8.39$ &\multicolumn{1}{c|}{$7.48\pm3.48$} & $9.63\pm7.27$ &  \multicolumn{1}{c|}{$7.76\pm4.60$} & $9.83\pm5.66$ &\multicolumn{1}{c|}{$4.72\pm4.35$} & $47.93\pm15.42$\\
    \multicolumn{1}{c|}{NL (identity)} & $5.20\pm3.49$ &\multicolumn{1}{c|}{$1.06\pm0.64$} & $5.60\pm2.94$ & \multicolumn{1}{c|}{$4.37\pm3.34$} & $11.49\pm6.98$ &\multicolumn{1}{c|}{$7.10\pm5.02$} & $64.44\pm4.68$ \\
    \multicolumn{1}{c|}{NL (isotonic)} & $5.10\pm3.77$ &\multicolumn{1}{c|}{$1.07\pm0.91$} & $6.28\pm4.26$ &\multicolumn{1}{c|}{$4.12\pm2.32$} & $7.46\pm5.35$ &\multicolumn{1}{c|}{$4.62\pm2.78$} & $46.80\pm16.83$ \\
    \multicolumn{1}{c|}{NL (isotonic$^{*}$)} & $4.16\pm3.77$  &\multicolumn{1}{c|}{\textbf{0.63 }$\pm$\textbf{ 0.50}} & \textbf{5.02 }$\pm$\textbf{ 4.11} &\multicolumn{1}{c|}{\textbf{4.00 }$\pm$\textbf{ 2.96}} & \textbf{6.26 }$\pm$\textbf{ 3.94} &\multicolumn{1}{c|}{\textbf{4.61 }$\pm$\textbf{ 3.08}} & $50.35\pm6.46$\\
    \multicolumn{1}{c|}{Population (identity)} & $5.35\pm2.74$  &\multicolumn{1}{c|}{$2.24\pm1.50$} & $9.89\pm8.18$ &\multicolumn{1}{c|}{$7.76\pm6.50$} & $8.28\pm5.62$ &\multicolumn{1}{c|}{$6.02\pm4.23$} & $45.69\pm7.10$\\
    \multicolumn{1}{c|}{Population (isotonic)} & $4.03\pm2.17$  &\multicolumn{1}{c|}{$1.27\pm1.00$} & $7.48\pm5.36$ &\multicolumn{1}{c|}{$6.66\pm4.11$} & $6.73\pm5.50$ &\multicolumn{1}{c|}{$4.75\pm4.17$} & $61.79\pm2.62$\\
    \multicolumn{1}{c|}{Population (isotonic$^{*}$)} & \textbf{3.74 }$\pm$\textbf{ 2.43}   &\multicolumn{1}{c|}{$0.80\pm0.60$} & $5.27\pm3.63$ &\multicolumn{1}{c|}{$4.30\pm2.74$} & $7.39\pm5.23$ &\multicolumn{1}{c|}{$4.65\pm3.04$} & $63.70\pm4.18$\\
    \Xhline{2\arrayrulewidth}
    \multicolumn{1}{c|}{Cost saved on images (\$)} & $163,692,910$  &\multicolumn{1}{c|}{$163,678,178$} & $22,906$ &\multicolumn{1}{c|}{$22,600$} & $2,459,183$ &\multicolumn{1}{c|}{$2,408,995$} & $163,692,910$ \\
    \multicolumn{1}{c|}{Time saved on labeling (hours)} & $1,871,956$  &\multicolumn{1}{c|}{$1,871,788$} & $266$ &\multicolumn{1}{c|}{$263$} & $28,123$ &\multicolumn{1}{c|}{$27,549$} & $1,871,956$ \\
    \Xhline{3\arrayrulewidth}
    \end{tabular}
    }
    \label{tab:object_count}
\end{table*}

\begin{figure*}
    \centering
    \includegraphics[width=1.0\textwidth]{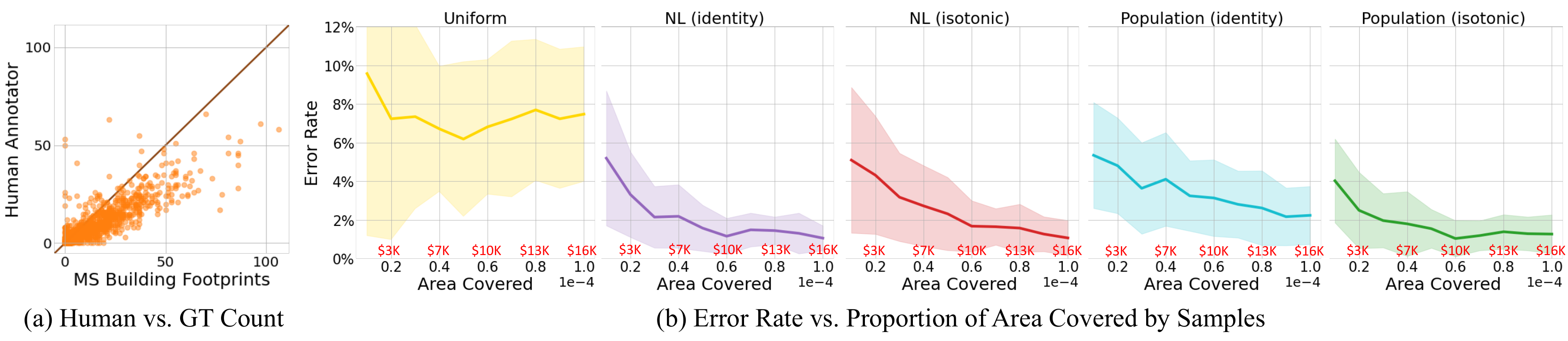}
    \caption{Figure (a): Counts from existing building datasets are consistent with counts from human annotators on the same images.
    Due to potential tendencies of excluding objects on the boundaries, human annotator can give slightly smaller counts than counts from the building datasets.
    Figure (b): Estimation error of building counts at country-level for the United States averaged over 20 runs. The total number of buildings in the US. from MS Building Footprints is treated as the ground truth (GT). We plot the ratio between the area covered by the used images and the area of the US. on the x-axis and label the corresponding cost for purchasing images (in USD). We observe that \model{} has smaller variances than uniform sampling and converges faster to the true building count than uniform sampling.
    }
    \label{fig:correlation}
\end{figure*}

\section{Results}
In this section, we evaluate the performance of \model{} on the tasks introduced in the previous section. 
We show with empirical results that \model{} drastically reduces the variance of sampling in most settings, leading to huge savings of up to \$163millions for purchasing high-resolution satellite images and 1 million hours for image labeling  with human annotators compared to an exhaustive approach. We provide extra details in Appendix.

\subsection{\model{} with Base Proposal Distributions}
To evaluate the performance of importance sampling with covariates as the proposal distributions, we compare \model{} (identity) with uniform sampling in this section.
In \Cref{tab:object_count}, we show the errors of object count estimation in different tasks, where  %
\model{} consistently outperforms uniform sampling by a large margin on counting buildings and cars. 
Moreover, as sample size increases, the estimates based on \model{} converge quickly while the estimates based on uniform sampling show no obvious trend of convergence (see \Cref{fig:correlation}(b)), whereas covariate-based estimates have reduced variance.

It is interesting to note that all methods give a high error rate on the count of swimming pools, and all estimates converge to approximately the same value (see the last column of \Cref{tab:object_count}). One plausible reason is that a significant number of swimming pools are indoors, and therefore not visible in satellite images. 
However we are not aware of data sources on the count of outdoor swimming pools to perform additional evaluation. Given that all approaches in \Cref{tab:object_count} converge to approximately the same counts and our method has strong performance on the other tasks, we believe \model{} should provide a reasonably accurate estimate of outdoor swimming pool counts.

The choice of covariates for building the proposal distribution also affects the estimation. On the car counting task (see \Cref{tab:object_count}),
NL (identity) method outperforms both uniform and population (identity) methods, while NL (identity) does not outperform the uniform sampling on brick kiln counting (see \Cref{tab:object_count}).
We believe one potential reason is that the distribution of cars could be more correlated with NL than population, 
while the distribution of brick kilns could  be more correlated with population than NL.
Therefore, in order to generate a task-specific proposal distribution with strong performance, we propose to fine-tune the covariate distribution using isotonic regression.

\subsection{\model{} with Tuned Proposal Distributions}
We observe that learning the proposal distribution with isotonic regression further improves the performance in most of the settings (\Cref{tab:object_count}).
In car and brick kiln experiments, we observe that fine-tuning with isotonic methods consistently improves the performance of identity methods even with the training size deducted from the sampling budget. 
In addition, in \Cref{fig:correlation}(b), we observe that the population (isotonic) proposal distribution converges fastest to the ground truth building count, compared to the population (identity) method. 
We believe our empirical results support the effectiveness of \model{}.

\subsection{Cost Analysis}
We compare the cost required for purchasing images for the exhaustive approach and \model{} in the last two rows of \Cref{tab:object_count}. We observe that \model{} saves as much as 99.99\% of the cost for purchasing images and 99.99\% hours needed for labeling with human annotators\footnote{Hour estimation based on Amazon Mechanical Turk}. When the target region is as large as the US,  \model{} saves \$163 million for purchasing images and 1 million hours for labelling images, while achieving less than 1\% error on building count estimation. We provide more details on cost estimation in Appendix.

\section{Discussion and Societal Impact}
Understanding where humans build things and what they build is a central component of measuring the productivity of economies and the livelihoods of individuals worldwide. %
Knowledge of this physical capital is also central for a range of policy decisions, from planning infrastructure investments to delivering services to adapting to climate change.  Finally, accurate measurement of the physical capital stock over time is central to answering fundamental questions about sustainable development, in particular in understanding whether we are leaving future generations as well or better off than current and past generations~\cite{solow1991sustainability}.

We provide a new approach to accurately estimate object counts at large scale and low cost.  
Compared to existing brute-force approaches which require purchasing vast amounts of high-resolution images at very high cost (\eg, \$164 million for the entire US), our approach enables high accuracy counts of objects while eliminating 99.99\% of the cost of purchasing and labeling imagery. As our \model{} model is scalable to large-scale object counting and can be easily adapted to different tasks, it is applicable to a broad range of research and policy tasks, including contributing to near-term tracking of multiple Sustainable Development Goals. For instance, the number of cars and buildings reflects economic development in low-income countries and can be used to measure livelihoods directly~\cite{Ayush2021EfficientPM} (SDG 1 No Poverty), and the number of brick kilns reflects pollution from informal industries (SDG 13 Climate Action).   Our approach can also make real-world counting efforts more inclusive for researchers and policymakers with more limited budgets, including those from developing countries, democratizing contributions and facilitating progress towards real-world sustainability applications.

\section*{Acknowledgements}
The authors would like to thank everyone from the Stanford Sustainability and AI Lab for the constructive feedback and discussion.
This work was supported by NSF awards (\#1651565, \#1522054), the Stanford Institute for Human-Centered AI (HAI), the Stanford King Center, the United States Agency for International Development (USAID), a Sloan Research Fellowship, and the Global Innovation Fund.
WN was supported in part by ONR (N000141912145), AFOSR (FA95501910024), ARO (W911NF-21-1-0125), DOE (DE-AC02-76SF00515).

\bibliography{aaai22}

\clearpage
\appendix

\section{Appendix A: Proofs}
\label{app:sec:proof}
\thm*
\paragraph{Proof}
Proposition 1 follows directly from Theorem 1.1 in \cite{chatterjee2018sample}. By letting $f(y)=C$, $\nu=q^{*}(\x)$, $\mu=q_{\theta}(\x)$ in \cite{chatterjee2018sample}, we obtain 
\begin{equation}
\resizebox{0.95\linewidth}{!}{
    $\mathbb{E}[\vert \hat{C}_n-C\vert ]\le C\bigg(e^{-\frac{t}{4}}+2\sqrt{\mathbb{P}(\log \frac{q^{*}(\x)}{q_{\theta}(\x)}>L+\frac{t}{2})}\bigg)$.
}    
\end{equation}

\inequality*
\paragraph{Proof}
Since $f(\x, l)$ is non-negative by definition, $\hat{C}_n$ is non-negative. According to Markov's inequality, we have 
\begin{equation}
    \mathbb{P}(\hat{C}_n\ge a)\le \frac{\mathbb{E}(\hat{C}_n)}{a}
\end{equation}
for any $a>0$.
Let $a=k\mathbb{E}(\hat{C}_n)=kC$ for some $k>0$, we have 
\begin{equation}
    \mathbb{P}(\hat{C}_n\ge kC)\le \frac{1}{k}.
\end{equation}

\variance*
\paragraph{Proof}
Denote $r = q^{*} / q$, which is the normalized ratio between $f(\x,l)$ (see \Cref{fig:counting_box}) and the proposal distribution $q(\x)$.
\begin{align}
    \mathrm{Var} & = C^2 \mathrm{Var}_q[r] \\
    & = C^2 \mathbb{E}_q\left[\left(\frac{q^{*}(\rvx)}{q(\rvx)}\right)^2\right] - C^2 \left(\mathbb{E}_q\left[\frac{q^{*}(\rvx)}{q(\rvx)}\right]\right)^2 \label{eq:vardef} \\
    & = C^2 (\mathbb{E}_{q^{*}}\left[\frac{q^{*}(\rvx)}{q(\rvx)}\right] - 1)  \label{eq:rdt} \\
    & \geq C^2 (e^{\mathbb{E}_{q^{*}}[\log \frac{q^{*}(\rvx)}{q(\rvx)}]} - 1) \label{eq:log-x}\\
    & = C^2 (e^{KL(q^{*} \Vert q)} - 1). \label{eq:kldef}
\end{align}

\section{Appendix B: Cost Computation Details}
We assume a per sq km price of \$17 for high-resolution satellite images used in this project, consistent with the current industry rates. To compute the cost on purchasing images given a region, we multiply the area of the region by the per sq km price. For example, the total area of the US. is 9,629,091 sq km (excluding the territories and dependencies)\footnote{\url{https://www.nationsencyclopedia.com/economies/Americas/United-States-of-America.html}}. Hence the total cost for collecting all images in the US is $9,629,091 \times \$17 \approx \$163,000,000$.

\section{Appendix C: Human Annotation Details}
\begin{figure}[H]
    \centering
    \includegraphics[width=1.0\linewidth]{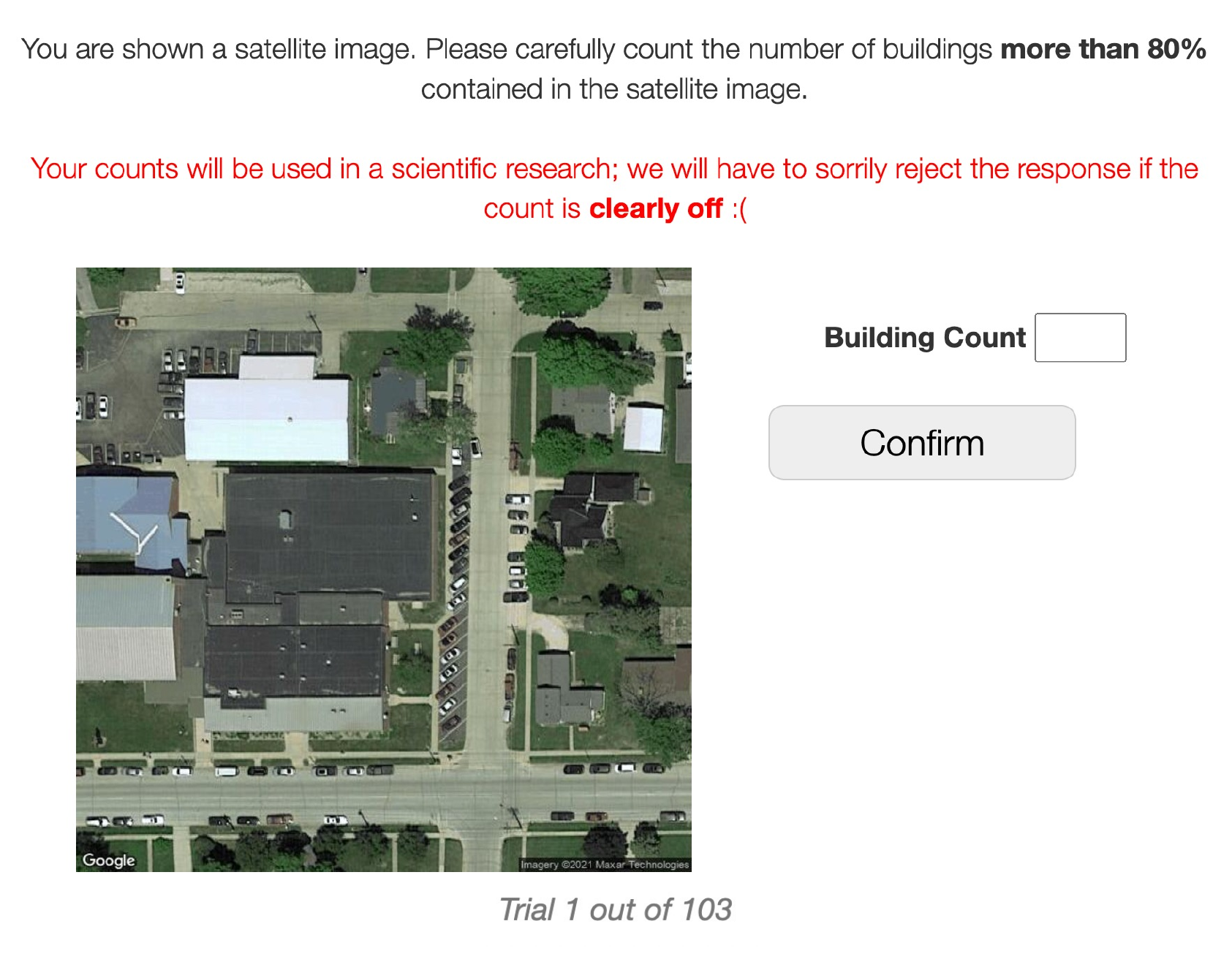}
    \caption{One example page of the MTurk HIT interface. The turker is asked to put the number of buildings more than 80\% contained in the image (as also noted on the page) into the text box. If the input is a negative or empty value, an alert will pop up.}
    \label{app:fig:mturk_interface}
\end{figure}

We use the Amazon Mechanical Turk (Mturk) platform for annotating images sampled using our methods. Each Human Intelligence Task (HIT) contains 103 images, including 3 vigilance images for ensuring the quality of the collected labels. To ensure a consistent definition of object boundaries across different human annotators to the greatest extent, we also provide practice trials along with examples of bounding boxes. We pay \$3 for each HIT. 
An example of the MTurk HIT interface is shown in \Cref{app:fig:mturk_interface}. 

According to the HIT statistics from MTurk, a turker can finish a HIT in 43 minutes on average. We treat this number as roughly the time needed to label 100 images of size $640\times640$pix and 0.3m GSD. 
For instance, to cover Kenya (total area 582,650 sq km)\footnote{\url{https://www.nationsencyclopedia.com/Africa/Kenya-LOCATION-SIZE-AND-EXTENT.html}} we need $582,650 / (640 \times 0.0003)^2 \approx 15,805,393$ images, and labeling all of them needs $(15,805,393 / 100 \times 43) / 60 \approx 115,000$ hours; to cover the US we need 261,205,810 images, and labeling all of them takes $(261,205,810 / 100 \times 43) / 60 \approx 1,000,000$ hours.

\begin{figure}[t]
    \centering
    \includegraphics[width=1.0\linewidth]{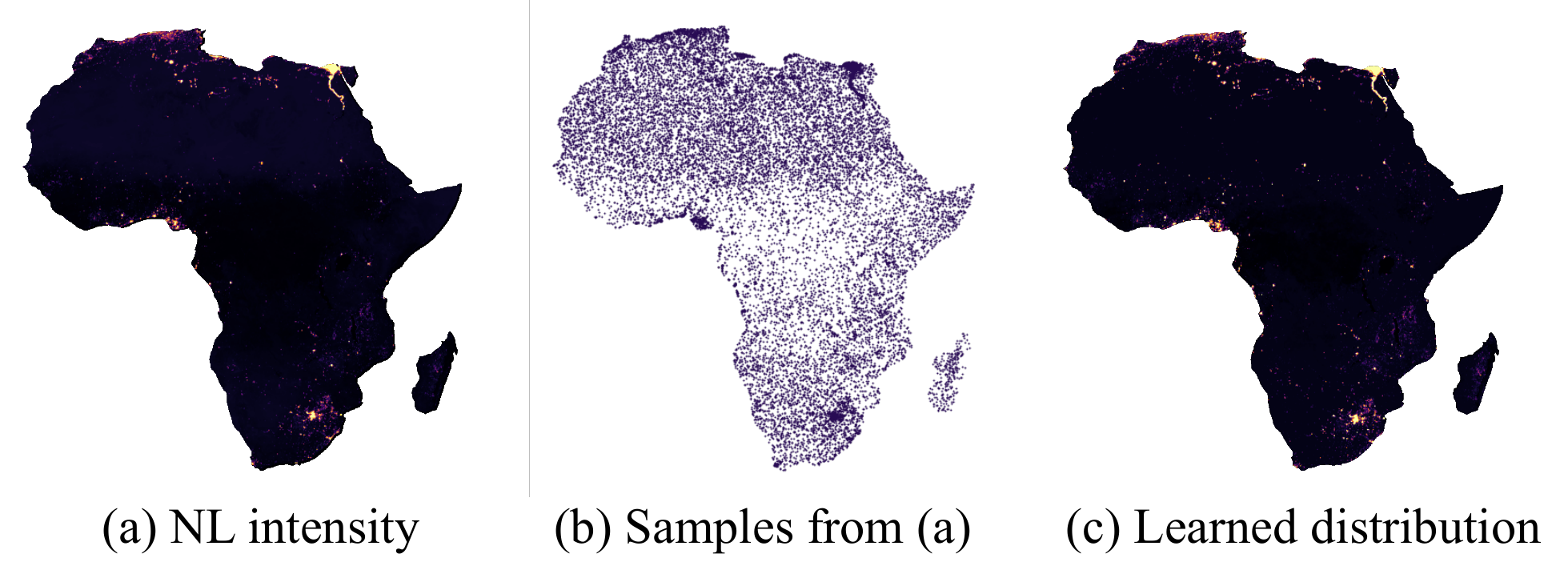}
    \caption{Visualization of distributions in Africa. Figure (a) shows the NL in Africa; Figure (b) shows samples drawn from the NL distribution in (a); Figure (c) shows the proposal distribution learned by isotonic regression with (a) as inputs.}
    \label{app:fig:nl_data}
\end{figure}

\begin{figure*}[h]
    \centering
    \includegraphics[width=1.0\textwidth]{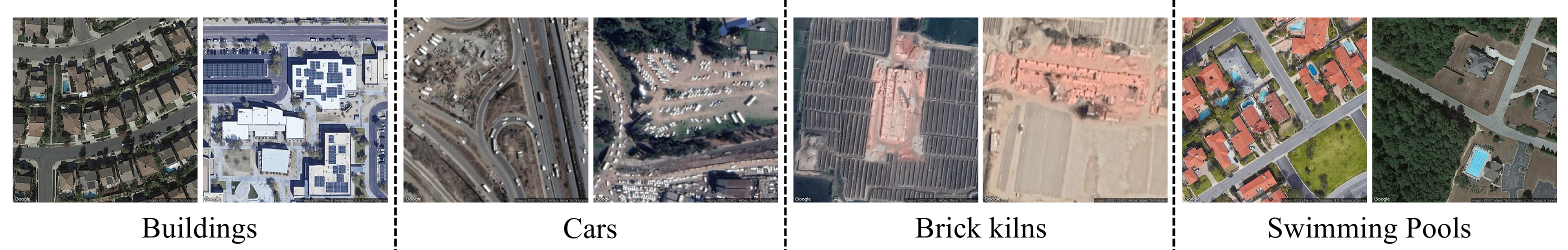}
    \caption{Examples of buildings, cars, brick kilns, and swimming pools in satellite images}
    \label{app:fig:object_examples}
\end{figure*}

\begin{figure*}[htp]
    \centering
    \includegraphics[width=1.0\textwidth]{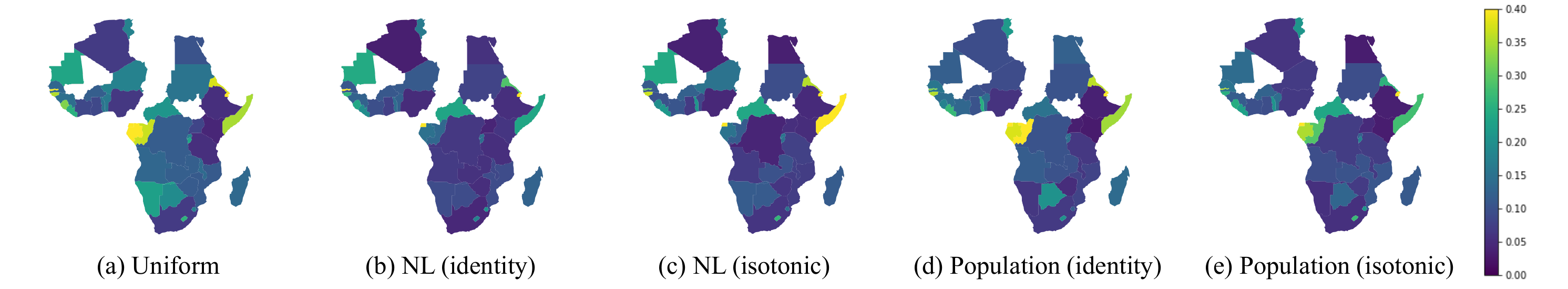}
    \caption{Estimation error on 43 African countries using $0.1\%$ images in each country (20-run average).}
    \label{app:fig:building_afr_error}
\end{figure*}

\section{Appendix D: Additional Experiment Details}
In this section we provide additional experiment details.

\subsection{Buildings}
For the experiment of counting buildings, use the Microsoft Building Footprints\footnote{\url{https://github.com/microsoft/USBuildingFootprints}} and the Google Open Buildings \cite{Sirko2021ContinentalScaleBD}
as the ground truth building distribution for the US. and Africa respectively. 
Both datasets provide polygon bounding boxes of the buildings labeled by models or human experts, and the Google Open Buildings further provides a confidence score of the prediction. 
Specifically, the MS Building Footprints dataset includes 130,099,920 footprints in the US. For the Google Open Buildings dataset, we only consider buildings with confidence scores higher than 80\% to avoid unreasonable predictions, yielding a total of 176,179,667 buildings in the 43 African countries. During experiment, the image-wise building counts are extracted as the number of polygon bounding boxes whose centroids fall within the images.

The error averaged over 51 US. states and 43 African countries using different methods are shown in \Cref{app:tab:object_count_state_averaged}. To show the effectiveness of isotonic regression, we also compare it with two monotonic transformations (see the last two rows of \Cref{app:tab:object_count_state_averaged}).

We also provide additional experiment results on counting buildings in 43 African countries and present the error visualization in \Cref{app:fig:building_afr_error}. The full list of 43 African countries is shown below:
\textit{Algeria, Angola, Benin, Botswana, Burkina Faso, Burundi, Central African Republic, Cote d Ivoire, Democratic Republic of the Congo, Djibouti, Egypt, Equatorial Guinea, Eritrea, Eswatini, Ethiopia, Gabon, Gambia, Ghana, Guinea-Bissau, Guinea, Kenya, Lesotho, Liberia, Madagascar, Malawi, Mauritania, Mozambique, Namibia, Niger, Nigeria, Republic of the Congo, Rwanda, Senegal, Sierra Leone, Somalia, South Africa, Sudan, Tanzania, Togo, Tunisia, Uganda, Zambia, Zimbabwe}

\subsection{Cars}
For the sake of validation on car count, we download and hand-label all image tiles in the rectangular region surrounding Nairobi, the capital of Kenya, covering approximately 1383km$^2$. There are in total $38,158$ non-overlapping and continuous image tiles of size $640\times640$pix and 0.3m GSD.

\subsection{Brick Kilns}
The dataset consists of 6978 brick kilns in Bangladesh from the period of October 2018 to May 2019 labeled by an exhaustive enumeration using a combination of ML models and human experts \cite{lee2021scalable}.

\begin{table}[t]
    \centering
    \caption{Averaged error over all states in the US. and 43 African countries. Results are averaged over 20 runs. The total area of used satellite images covers 0.1\% of each target state (US) and country (Africa). The results using Uniform (identity) method is provided in the parenthesis following the region name. For \textbf{isotonic}, training size (20\% of the sample budget) is deducted from the sampling budget. For \textbf{isotonic$^*$}, training size (20\% of the sample budget) is not deducted from the sampling budget. 
    }
    \resizebox{\linewidth}{!}{
    \begin{tabular}{cccccccc}
    \Xhline{3\arrayrulewidth}
    &\multicolumn{2}{c}{US states (16.77\%)}
    &\multicolumn{2}{c}{African countries (21.02\%)}\\
    \Xhline{2\arrayrulewidth}
    Methods &\multicolumn{1}{c}{NL}
    &\multicolumn{1}{c}{Population}
    &\multicolumn{1}{c}{NL}
    &\multicolumn{1}{c}{Population}\\
     \Xhline{2\arrayrulewidth}
     Identity &8.46\% &9.71\%  %
     &15.58\% &18.27\% \\
     Isotonic &8.80\% &8.08\% &15.79\% &17.43\%\\
     Isotonic$^{*}$ &8.09\% &7.48\% &14.86\% &16.85\%\\
     $g(\x, l)=e^{\x}$ &16.83\% &16.23\% &21.04\% & 21.75\% \\
     $g(\x, l)=\log(\x)$ &18.01\% &21.47\% &22.98\% &23.39\%\\
    \Xhline{3\arrayrulewidth}
    \end{tabular}
    }
    \label{app:tab:object_count_state_averaged}
\end{table}

\subsection{Remote-Sensing Data}
For all experiments, we use satellite images of size $640\times640$pix and 0.3m GSD.
Examples of the four object categories (\ie buildings, cars, brick kilns, and swimming pools) in satellite images are shown in \Cref{app:fig:object_examples}.

\subsection{Covariate Data}
We use raster data for global nightlight and population density. Specifically, we used the 2012 gray scale nightlight raster with resolution 0.1 degree per pixel downloaded from NASA\footnote{\url{https://earthobservatory.nasa.gov/features/NightLights}}. 
For population density, we downloaded the 2020 Gridded Population of the World, Version 4, with a resolution of 30 seconds per pixel\footnote{\url{https://sedac.ciesin.columbia.edu/data/set/gpw-v4-population-density-rev11}}. 

We preprocess the raster data to construct the base distribution. Given a region of interest, we first collect publicly available region boundary shapefile\footnote{\url{https://github.com/nvkelso/natural-earth-vector}}. Then we compute the pixel values of all pixels inside the region boundary. The pixel value is computed as the average RGB value of that pixel normalized over all target pixels that fall into the region. Given region $R$, the pixel value $Val(x_i)$ of a pixel $x_i$ is computed as
\begin{equation*}
    Val(x_i) = \frac{RGB(x_i)}{\sum_{x \in R} RGB(x)}
\end{equation*}
where $RGB(\cdot)$ indicates the average RGB value of the pixel. In this way, we build the base distribution of the target region from the covariate.

\subsection{Training Isotonic Regression}
The training data used to fine-tune the isotonic regression model is consist of 50\% positive samples (\ie with non-zero object count) and 50\% negative samples drawn from the uniform distribution. Note that only for swimming pools, the 50\% positive samples are drawn from the base distributions (\ie could have zero object count) because samples with non-zero counts are insufficient. 
We also cap the number of training samples at 5000 for buildings, brick kilns, and swimming pools to avoid spending too much on fine-tuning. 
For cars, considering the limited amount of labeled data, we cap the number of training samples at 200 instead.

\section{Appendix E: YOLOv3-based Regression Model Training Details}

We train a YOLO-based regression model for counting the number of buildings using the xView dataset \cite{lam2018xview}, which provides high-resolution satellite images of $0.3$m GSD.

\subsection{Data Preprocessing}
We preprocess the xView dataset \cite{lam2018xview} into image tiles of size $416\times416$pix (\ie the input size of YOLO-v3). The original dataset provides object categories and the corresponding bounding boxes within each tile. For the purpose of training a model for predicting object counts, we use the number of bounding boxes with category ``Building'' as the label for each image. There are in total 36,996 image tiles, and we use 30,000 for training and the rest for testing.

\subsection{Architecture}
The YOLO-based regression model is composed of two parts: a feature extractor and a debiasing module. The feature extractor part uses the YOLO-v3 weights pretrained on ImageNet. Then, we replace the final detection layer of a YOLO-v3 model with a three-layer Multi-Layer Perceptron (MLP), which serves as a debiasing module for predicting the object count. The entire regression model is then fine-tuned on 30K xView images for more than 200K epochs until convergence. 

\section{Appendix F: Ablation Study on the Resolution of Covariate Raster}

While using a covariate raster with higher-resolution could intuitively give a better initial proposal distribution,
we observe empirically that improving the resolution of the rasters does not necessarily improve the estimation performance when the raster resolution is high enough.
For instance, we observe that the estimation error is $4.66\pm2.76$ (\%) for nightlight raster with 0.75km ground sampling distance (GSD),
$3.89\pm2.44$ (\%) for a lower-resolution raster with 3km GSD, and $4.02\pm3.14$ (\%) 
for a coarse raster with 11.1km GSD in New York State using samples covering 0.1\% of the total area.
As \model{} yields an unbiased estimator, the estimation will converge to the right count given sufficient samples despite the raster resolutions.

\end{document}